\documentclass[sigconf]{acmart}

\AtBeginDocument{%
  \providecommand\BibTeX{{%
    \normalfont B\kern-0.5em{\scshape i\kern-0.25em b}\kern-0.8em\TeX}}}

\usepackage{booktabs} 
\usepackage{makecell}
\usepackage[ruled, linesnumbered]{algorithm2e}

\usepackage{url}
\usepackage{latexsym}
\usepackage{amsmath}
\usepackage{graphicx}
\usepackage{graphics}
\usepackage{pifont}

\usepackage{lipsum}

\usepackage[subfigure]{graphfig}
\usepackage{multirow}
\usepackage{CJKutf8}
 
\usepackage[skip=0.5\baselineskip]{caption}
\usepackage{balance}
\usepackage{multirow, booktabs}
\usepackage{footnote}
\usepackage{listings}
\usepackage{csquotes}

\definecolor{codegreen}{rgb}{0,0.6,0}
\definecolor{codegray}{rgb}{0.5,0.5,0.5}
\definecolor{codepurple}{rgb}{0.58,0,0.82}
\definecolor{backcolour}{rgb}{0.95,0.95,0.92}

\lstdefinestyle{mystyle}{
    backgroundcolor=\color{backcolour},   
    commentstyle=\color{codegreen},
    keywordstyle=\color{magenta},
    numberstyle=\tiny\color{codegray},
    stringstyle=\color{codepurple},
    basicstyle=\ttfamily\footnotesize,
    breakatwhitespace=false,         
    breaklines=true,                 
    captionpos=b,                    
    keepspaces=true,                 
    numbers=left,                    
    numbersep=5pt,                  
    showspaces=false,                
    showstringspaces=false,
    showtabs=false,                  
    tabsize=2
}

\def\runningfoot{\def\@runningfoot{}}
\def\firstfoot{\def\@firstfoot{}}
\setcopyright{none} 

\begin{document}
\fancyhead{}

\title{Marrying Dialogue Systems with Data Visualization: \\ Interactive Data Visualization Generation from Natural Language Conversations}

\author{Yuanfeng Song}
 \affiliation{
   \institution{The Hong Kong University of Science and Technology \& WeBank Co., Ltd}
   \city{Hong Kong}
   \country{China}
 }
 \email{songyf@cse.ust.hk}

 \author{Xuefang Zhao}
 \affiliation{%
   \institution{AI Group \\ WeBank Co., Ltd}
   \city{Shenzhen}
   \country{China}
 }
 \email{summerzhao@webank.com}

 \author{Raymond Chi-Wing Wong}
 \affiliation{%
   \institution{The Hong Kong University of Science and Technology}
   \city{Hong Kong}
   \country{China}
 }
 \email{raywong@cse.ust.hk}

\begin{abstract}
Data visualization (DV) has become the prevailing tool in the market due to its effectiveness into illustrating insights in vast amounts of data.
To lower the barrier of using DVs, automatic DV tasks, such as natural language question (NLQ) to visualization translation (formally called \emph{text-to-vis}), have been investigated in the research community. However, text-to-vis assumes the NLQ to be well-organized and expressed in a \emph{single sentence}. However, in real-world settings, complex DV is needed through \emph{consecutive exchanges} between the DV system and the users. In this paper, we propose a new task named \textbf{CoVis}, short for \underline{Co}nversational text-to-\underline{Vis}ualization, aiming at constructing DVs through a series of interactions between users and the system. 
Since it is the task which has not been studied in the literature, we first build a benchmark dataset named \textbf{Dial-NVBench}, including dialogue sessions with a sequence of queries from a user and responses from the system. The ultimate goal of each dialogue session is to create a suitable DV. However, this process can contain diverse dialogue queries, such as seeking information about the dataset, manipulating parts of the data, and visualizing the data. Then, we propose a multi-modal neural network named \textbf{MMCoVisNet} to answer these DV-related queries. In particular, MMCoVisNet first fully understands the dialogue context and determines the corresponding responses. Then, it uses adaptive decoders to provide the appropriate replies: (i) a \emph{straightforward text decoder} is used to produce general responses, (ii) an \emph{SQL-form decoder} is applied to synthesize data querying responses, and (iii) a \emph{DV-form decoder} tries to construct the appropriate DVs. We comparatively evaluate MMCoVisNet with other baselines over our proposed benchmark dataset. Experimental results validate that MMCoVisNet performs better than existing baselines and achieves a state-of-the-art performance. 
\end{abstract}

\ccsdesc[500]{Human-centered computing~Visualization}
\ccsdesc[500]{Computing methodologies~Natural language generation}

\keywords{Data Visualization, Conversational Text-to-Visualization, Multi-modal Dialogue System, NLP for Data Mining}

\maketitle

\section{Introduction}
\label{sec:introduction}

\begin{displayquote}
\emph{``The future of BI (Business Intelligence) is Conversational! ''} 
-- Gartner Analyst Report
\end{displayquote}

Data visualization (DV) has emerged as the prevailing tool in the industry in summarizing the insight behind raw data. 
Given a massive dataset, DV could vividly and efficiently convey the essence behind the raw data than a verbal description \cite{iliinsky2011designing}.
Due to its great usage, DV has been widely adopted and explored by many commercial vendors (e.g., ThoughtSpot \cite{ts} and Microsoft Power BI \cite{ms}), as well as academic researchers \cite{srinivasan2017natural,song2022rgvisnet,dibia2019data2vis,narechania2020nl4dv}. 

To lower the barrier of using DVs, automatic DV tasks, such as \emph{text-to-vis} (i.e., natural language question (NLQ) to visualization translation), have been extensively explored in the data mining community \cite{luo2021synthesizing,luo2021natural,song2022rgvisnet}. 
For example, Song \textit{et al.} \cite{song2022rgvisnet} published a novel text-to-vis model in KDD'22 that employs a hybrid retrieval and generation framework towards accurate DV generation. 
However, all these existing studies expect the input NLQ to be well-organized and expressed in a \emph{single sentence} for existing text-to-vis methods. 
In real-world settings, complex DV is needed through \emph{successive exchanges} between the DV system and the users. 
Under this scenario, \emph{multi-turn interactions} between users and the DV system are frequently required to (1) clarify any ambiguous questions, (2) query the information about the dataset, (3) visualize the data, and (4) notify users about the returned data or charts. 

\begin{figure}[t!]
    \centering
    \includegraphics[width=0.48\textwidth]{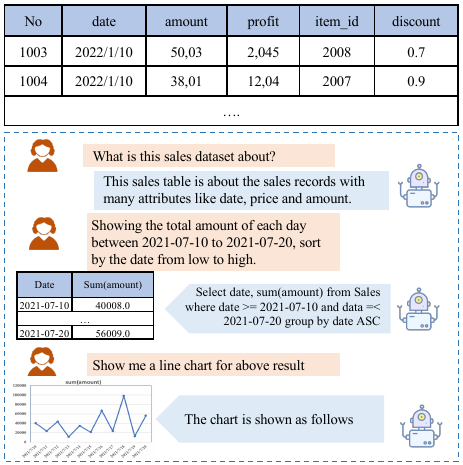}
    \vspace{-10pt}
    \caption{A conversation between a user and the data visualization dialogue system. The analyst gradually expresses his/her data querying and visualization needs gradually when the conversation progresses. Then, the dialogue system generates different responses according to the context, including text, data and DV Charts.}
    \label{fig:example}
    \vspace{-15pt}
\end{figure}

In this paper, we propose a new task named \textbf{CoVis}, short for \underline{Co}nversational text-to-\underline{Vis}ualization, aiming to construct DVs via a series of interactions between the system and the user. An illustrative example can be found in Figure~\ref{fig:example}, which shows that the analyst expresses his data querying and visualization needs gradually when the conversation progresses. The dialogue system generates different responses according to the context, including text, data (by SQL queries) and DV (by Vega-Lite \cite{satyanarayan2016vega}, one popular programming language dedicated for DV). 
The ultimate goal of each dialogue session is to create a suitable DV. However, this process can contain diverse user dialogue queries, like asking the information about the dataset, querying statistical information about the data, and visualizing parts of the data. 

Since CoVis has not been studied in the literature, we first construct a benchmark dataset, including a sequence of dialogues with queries and responses. The dataset is named \textbf{Dial-NVBench}, modified from NVBench \cite{luo2021synthesizing}, a text-to-vis dataset that contains single-turn (NLQ, DV) pairs. Since each sample in NVBench could be considered as a single-turn dialogue between the user and the chatbot, we greatly extend each sample from the NVBench dataset by adding other queries and responses aside from only querying about the desired DVs. Specifically, we add queries like greetings, asking for the details of the dataset, and the general response.

To construct a conversational text-to-vis system, we propose a multi-modal conversational network to answer these DV-related queries, and we name it \textbf{MMCoVisNet}. 
In particular, MMCoVisNet first fully understands the dialogue context and determines the corresponding responses. 
Then, it uses adaptive decoders to provide the appropriate replies: (i) a \emph{straightforward text decoder} is used to produce general responses, (ii) an \emph{SQL-form decoder} is applied to synthesize data querying responses, and (iii) a \emph{DV-form decoder} tries to construct the appropriate DVs.
We comparatively evaluate MMCoVisNet with other baselines over our proposed benchmark dataset. Extensive experiments validate that our method could considerably improve the baseline models. 

To sum up, the main contributions of this paper are as follows.
\begin{itemize}
\item We are the first to propose the CoVis task, which focuses on interactively constructing DVs given the massive dataset. 
CoVis could help people better comprehend the data and master how to create suitable DVs. 
We believe that this new CoVis task will enlarge the family of intelligent dialogue and DV systems. 

\item To validate the rationale of this new task, we construct a benchmark dataset {Dial-NVBench}, which could also promote the development of this field. Each dialogue session contains queries like querying the information about the dataset, querying parts of the data, and visualizing the data. 

\item We design the first multi-modal conversational DV network for this CoVis task - MMCoVisNet. MMCoVisNet is equipped with advanced encoders and adaptive decoders, which guarantees the final inference performance of the whole network. 

\item We conducted extensive experiments on our proposed dataset and compared it with other popular baselines. The results show that our proposed MMCoVisNet can perform better than several  strong counterparts. 
\end{itemize}

\begin{figure}[t!]
    \centering
    \includegraphics[width=0.50\textwidth]{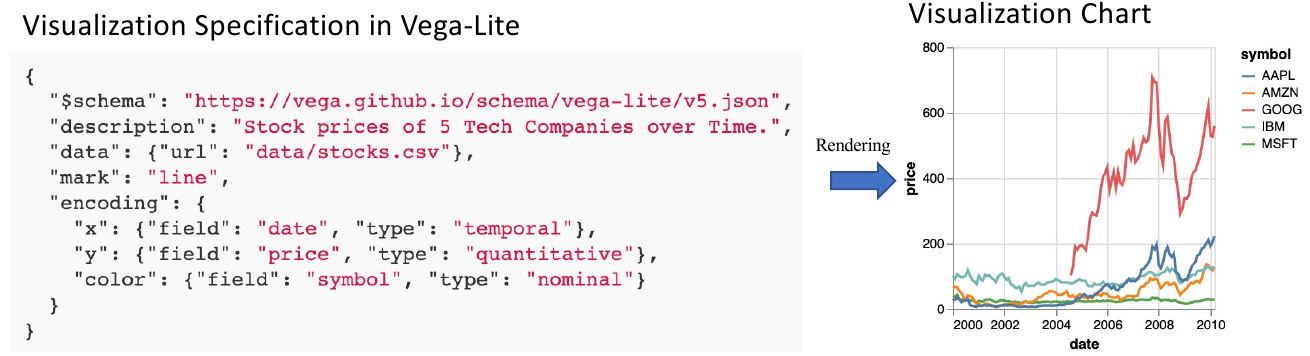}
    \caption{An example of the visualization specification in Vega-Lite, and its corresponding DV chart.}
    \label{fig:dv}
    \vspace{-15pt}
\end{figure}

\begin{figure*}[t!]
    \centering
    \includegraphics[width=1.0\textwidth]{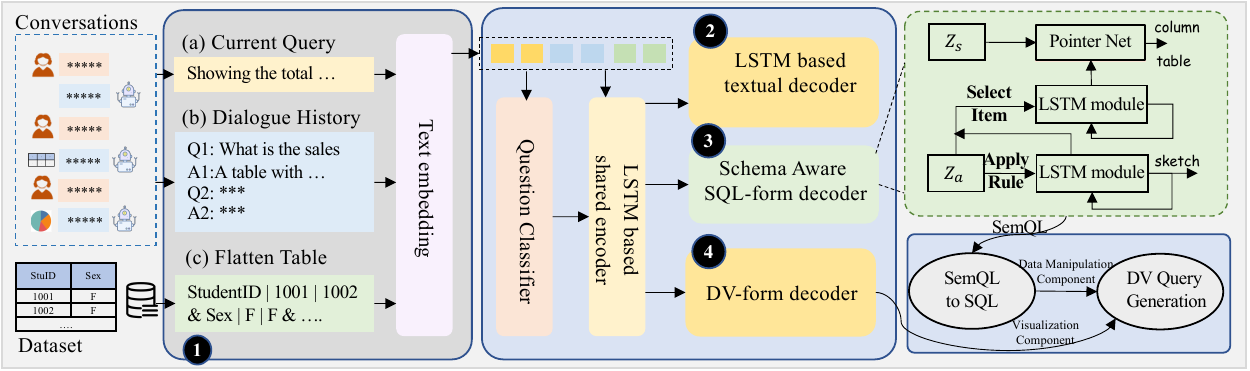}
    \vspace{-10pt}
    \caption{The network structure of the MMCoVisNet model, which is comprised of \ding{182} 
    a Data-aware Dialogue Session Encoder, 
    \ding{183} a Textual Decoder, 
    \ding{184} a SQL-form Decoder, and 
    \ding{185} a DV-form Decoder.
    To answer the data-related query, MMCoVisNet first generates SemQL via the SQL-form Decoder, then converts the SemQL into an SQL query, and finally obtains the data by executing the SQL query. Similarly, to answer the DV-related queries, MMCoVisNet first generates DV queries via the DV-form Decoder, then converts the DV queries to get the Vega-Lite specifications, and finally renders the specifications to get the charts.
    }
    \label{fig:network}
    \vspace{-10pt}
\end{figure*}

The rest of this paper is structured as follows: we first introduce basic concepts and the problem definition in Section~\ref{sec:prob}. 
Then, we discuss the details of our proposed MMCoVisNet model in Section~\ref{sec:model}. 
The experimental setup and results are listed in Section~\ref{sec:setup}, followed by the related work in Section~\ref{sec:related}. 
Finally, we conclude the work in Section~\ref{sec:con}.

\section{Concepts and Problem Definition}
\label{sec:prob}

In this section, we first provide some foundational concepts that help with the understanding of the following work, and these concepts can be divided into two categories, DV and CoVis. 

\subsection{Concepts on Data Visualization}

\noindent \emph{Declarative Visualization Language (DVL).} A DVL specifies the details of the visualization construction (e.g., chart type, color, size, mapping function, and properties for marks such as canvas size and legend). The most common DVLs include Vega-Lite \cite{satyanarayan2016vega}, ggplot2 \cite{villanueva2019ggplot2}, ZQL \cite{siddiqui2016effortless}, ECharts \cite{li2018echarts} and VizQL \cite{hanrahan2006vizql}, each of which has its own grammar. Our work mainly uses Vega-Lite due to its popularity and wide usage \cite{luo2021synthesizing, song2022rgvisnet, luo2021natural, tang2022sevi}. However, the methodology proposed can easily be applied to other DVLs.

\noindent \emph{Visualization Specification.} A visualization specification defines the exact properties of a DV using a DVL. In Figure~\ref{fig:dv}\footnote{More complicated examples, including this one, can be found in the official website of Vega-Lite via \textit{\url{https://vega.github.io/vega-lite/examples/}}.}, the specification in Vega-Lite is a JSON file specifying properties such as data path, mark and encoding. 
A specification could be rather complicated, and creating a suitable specification for a given dataset requires experience in DVL as well as familiarity with the raw data. 

\noindent \emph{DV Query.} To abstract all the DVLs, a inter-media form named DV query is proposed and widely used in the DV tasks \cite{luo2021synthesizing,luo2021natural,tang2022sevi,song2022rgvisnet}. 
Existing text-to-vis studies first synthesizes the DV query from the NLQ, and then executes the DV query to obtain the Vega-Lite specification. 
DV query enjoys a SQL-alike grammar with \emph{visualization component} and \emph{data manipulation component}. 
The final chart is obtained by rendering the specification file via the visualization engine \cite{satyanarayan2016vega}. 

\subsection{CoVis Problem Definition}
The definition of the \emph{CoVis} (Conversational Text-to-Visualization) task is relatively straightforward once the previous concepts are defined.
Specifically, given a dataset $D$, a query $q$ from the user with its context $S$, the CoVis task aims to predict the response $r$ automatically. 
CoVis enables users to create suitable DVs interactively. 
To fulfill the above objective, the context $S$ includes all the previous queries and responses in the same session and they are usually in the DV domain.
The query $q$ can be a diverse dialogue query, like seeking the information about the dataset, manipulating parts of the data, and visualizing the data. 
A desired CoVis model could be represented as $f(D, S, q) \rightarrow r$. 
Then, the complete training set can be represented as $\mathcal{T} = \{D^{(o)}, S^{(o)}, q^{(o)}, {r}^{(o)}\}^{N}_{o=1}$, where $N$ is the dataset size. 

\begin{table}[t!]
\caption{The Statistics of the proposed {Dial-NVBench} Dataset}
\centering
  \begin{tabular}{l|cccc}
    \toprule
    \textbf{Attributes} & \textbf{Train} & \textbf{Val.} & \textbf{Test} & \textbf{Total}  \\ 
    \midrule
    No. of Dialogue Sessions & 3115 & 484 & 896 & 4495 \\
    No. of QRs & 15044 & 2359 & 4332 & 21725 \\
    Avg. No. of QRs per Session & 4.83 & 4.874 & 4.824 & 4.833 \\
    No. of Datasets & 137 & 69 & 107 & 143 \\
    No. of General Queries & 6197 & 977 & 1797 & 8971 \\
    No. of Data-related Queries & 5732 & 898 & 1629 & 8259 \\
    No. of DV-related Queries & 3115 & 484 & 896 & 4495 \\
    \bottomrule
  \end{tabular}
\label{tab:dataset}
\vspace{-10pt}
\end{table}

\subsection{CoVis Dataset Construction}
Since CoVis is a new kind of DV-related dialogue tasks, there is currently no literature or public datasets for it. 
Thus, we first construct and release a benchmark dataset to promote the development of this field.
In particular, we synthesize the CoVis dataset by piggybacking the \textit{NVBench} dataset \cite{luo2021synthesizing} from the correlated text-to-vis task. 
Each sample in the NVBench contains a \emph{single-turn} NLQ with the corresponding DV (in terms of a DV query). 
We convert each of them into a \emph{dialogue session} by 
(i) randomly deciding the number of rounds (or \emph{query-responses (QRs)}) for the dialogue; 
(ii) sampling from some pre-defined general queries about the statistics of the dataset; 
(iii) generating some SQL-related queries about data manipulation, including ones involving complicated data operations like nested queries and joins, with the method similar to \cite{yu2021grappa};
and (iv) ending each session with a query (the same as the one contained in NVBench) about a DV. 
To validate this CoVis task, in this new dataset, we mainly focus on data querying and visualization on single tables. 
Finally, we obtained a dialogue dataset dedicated to DV, and we named it \emph{Dial-NVBench} (``Dial'' is short for ``Dialogue''). 

\section{Our Proposed Approach}
\label{sec:model}
In this section, we mainly discuss the proposed MMCoVisNet model from its main components. 

\vspace{-5pt}
\subsection{Model Overview}
As shown in Figure~\ref{fig:network}, MMCoVisNet mainly comprises four components, namely a \textit{Dialogue Session Encoder}, a \textit{Textual Decoder}, a \textit{SQL decoder}, and a \textit{DV Decoder}.
The dialogue session encoder takes the dataset, dialogue history and current query as input and then maps them into some hidden representation. 
The dialogue history is essential in the CoVis scenario since it helps identify item references in multi-turn dialogue sessions. 
If the desired response belongs to the general textual category, the textual decoder is used to generate a typical textual response, similar to existing dialogue systems. However, if the query is about manipulating the data, then the SQL decoder is triggered to generate SQL queries to obtain the data.  
To increase the conversion accuracy, MMCoVisNet employs SemQL \cite{guo2019towards} as an inter-media form for each SQL query. 
Similarly, MMCoVisNet uses a DV decoder to obtain the final charts to answer DV-related queries. 
Following the common practice in text-to-vis \cite{luo2021synthesizing,luo2021natural,tang2022sevi,song2022rgvisnet}, MMCoVisNet also first synthesizes the intermedia form - the DV query, and then executes the DV query to obtain the Vega-Lite specification. 
The final chart is obtained by rendering the specification via the visualization engine \cite{satyanarayan2016vega}. 

\subsection{Data-aware Dialogue Session Encoder}
\label{subsec:dataAwareDialogueSessionEncoder}

The objective of the dialogue session encoder is to convert the current query $q$, incorporated with a dialogue history and the dataset information, into a hidden representation. 
Given the dialogue history, we first concatenate them by a separate character `|', and then consider it a textual sequence. The whole process can be denoted as:
\begin{equation}
\begin{aligned}
s_{d}=d_1 | d_2 | \cdots |d_n.
\end{aligned}
\end{equation}
where $d_i$ is the sequence of the $i$-th dialogue session and $n$ is the total number of the dialogue sessions.
Since the dataset information dramatically affects the desired response, we must also preserve the schema information. 
Moreover, we sample some data records from the dataset as a snapshot for this dataset. 
To restrict the input sequence length, a certain number of rows are first selected randomly, and after we retain all the names of the schema items, we add the table values from the selected rows word by word until we reach the limit. Finally, the dataset sequence is formatted as
\begin{equation}
\begin{aligned}
s_{t}= c_1 | v_{11} | v_{12} | ... \& c_2 | v_{21} | ... ,
\end{aligned}
\end{equation}
where $c_i$ is the name of the column $i$, $v_{ij}$ is the $j$-th value of column $c_i$, `|' and `\&' are the separate characters for the items and rows, respectively.

The input sequence $Q$ of this session encoder is composed of the current query $q$, the history sequence $s_{d}$, and the dataset sequence $s_{t}$. 
We first map each token in $Q$ into its embedding (all these embeddings forms $\mathbf{Q}$) by a pre-trained Glove \cite{pennington2014glove} and then a bi-directional LSTM network (BiLSTM) \cite{yu2019review} is employed to deal with the input embedding to obtain the hidden vectors for the current input. The whole process is represented as 
\begin{equation}
\begin{aligned}
\mathbf{h}^Q_{t}=\texttt{\textup{BiLSTM}}(\mathbf{Q}_t, {\mathbf{h}}^Q_{t-1}),
\end{aligned}
\end{equation}
where $\mathbf{Q}_t$ is the item from $\mathbf{Q}$ at time step $t$ 
and ${\mathbf{h}}^Q_{t}$ represents the hidden states at time step $t$ for BiLSTM. 
Finally, we obtain the output $\boldsymbol{H}_Q = \{ \mathbf{h}^Q_1, \mathbf{h}^Q_2\, ..., \mathbf{h}^Q_{l} \}$ from the session encoder, where $l$ is the number of tokens in sequence $Q$.

\subsection{Adaptive Response Decoder}

Since the output of the CoVis task is multi-modal, we design an adaptive decoder to generate diverse responses. The adaptive decoder first uses a classifier to judge the desired response type. Then, three specific decoders are proposed for the downstream process: (i) a textual one for the textual output, (ii) an SQL-form one for the data query, and (iii) a DV-form one for the visualization query.

\subsubsection{General Textual Decoder}
\label{sec: textual_decoder}
The objective of this decoder is to generate textual sequences to answer questions about basic information. A Transformer-based decoder that takes the advantage of a multi-head scaled dot-product attention \cite{vaswani2017attention} is employed here, where multi-head attention can be formulated as the following equation. Specifically, given the query  $\hat{Q}$, the key $\hat{K}$, and the value $\hat{V}$, they are first injected into different high-dimensional spaces \footnote{The query $\hat{Q}$ in the attention mechanism is different from the previously defined $Q$ in CoVis, and thus, we use two different variables.}. 
Then, they perform an attention calculation to obtain a hidden representation for each head, and finally combine them to produce a final attention output.
\vspace{-5pt}
\begin{equation}
\begin{aligned}
head_i = \texttt{Attention}(\hat{Q}W^{\hat{Q}}_{i}, \hat{K}W^{\hat{K}}_{i}, \hat{V}W^{\hat{V}}_{i}),
\end{aligned}
\end{equation}
\vspace{-5pt}
\begin{equation}
\begin{aligned}
\texttt{MultiHead}(\hat{Q}, \hat{K}, \hat{V}) = [head_1; \cdots; head_g]W^{O},
\end{aligned}
\end{equation}
where $W^{\hat{Q}}_{i}, W^{\hat{K}}_{i} , W^{\hat{V}}_{i} \in \mathbb{R}^{{d_{k} \times d_\textup{m}} }$ and $W^{O} \in \mathbb{R}^{d_{m} \times d_{\textup{m}}}$ are trainable parameter matrices.
$d_k$ is the dimensionality of the input key vector $\hat{K}$, $d_{m}$ is the hidden size of the model and $g$ is the number of the heads.
The attention process consists of the similarity score calculation obtained by a dot-product operation between $\hat{Q}$ and $\hat{K}$ and a re-assignment according to the similarity score and the given $\hat{V}$, as shown in Equation \ref{eq:attention}.
\vspace{-5pt}
\begin{equation}
\begin{aligned}
\texttt{Attention}(\hat{Q}, \hat{K}, \hat{V}) = \texttt{Softmax}(\frac{\hat{Q}\hat{K}^{T}}{\sqrt{d_k}})\hat{V},
\end{aligned}
\label{eq:attention}
\end{equation}

Furthermore, the decoder is stacked with $K$ identical blocks composed of a self-attention layer, a cross-attention layer, and a feed-forward layer. Both self-attention and cross-attention follow the process of the aforementioned multi-head attention but have different inputs and objectives. 
We use $Z^k_i$ and $Z^k_o$ to denote the input and output for the $k$-th block ($0\leq k\leq K$), and $Z^k_s$ to denote the output of self-attention.
The word embedding $R$ for the target textual response $r$ is set as the initial input (i.e., $Z^0_o = R$).
Since it contains multiple layers, the output of previous one is set as the input for the following one (i.e., $Z_i^{(k)} = Z_o^{(k-1)}$, $\hat{Q}=\hat{K}=\hat{V}=Z^k_i$). 
It thus makes sure that the target textual response can go through and pay attention to itself and finally generate $ Z_s^{(k)}$.
In contrast, the cross-attention component takes the outputs from both the self-attention layer and the encoder as the inputs (i.e., $\hat{Q}=Z^k_{s}, \hat{K}=\hat{V}=\boldsymbol{H}_Q$) to produce a representation for each target sequence word that captures the influence from the input sequence.

\begin{figure}[t!]
\centering
    \centering
    \includegraphics[width=0.43\textwidth]{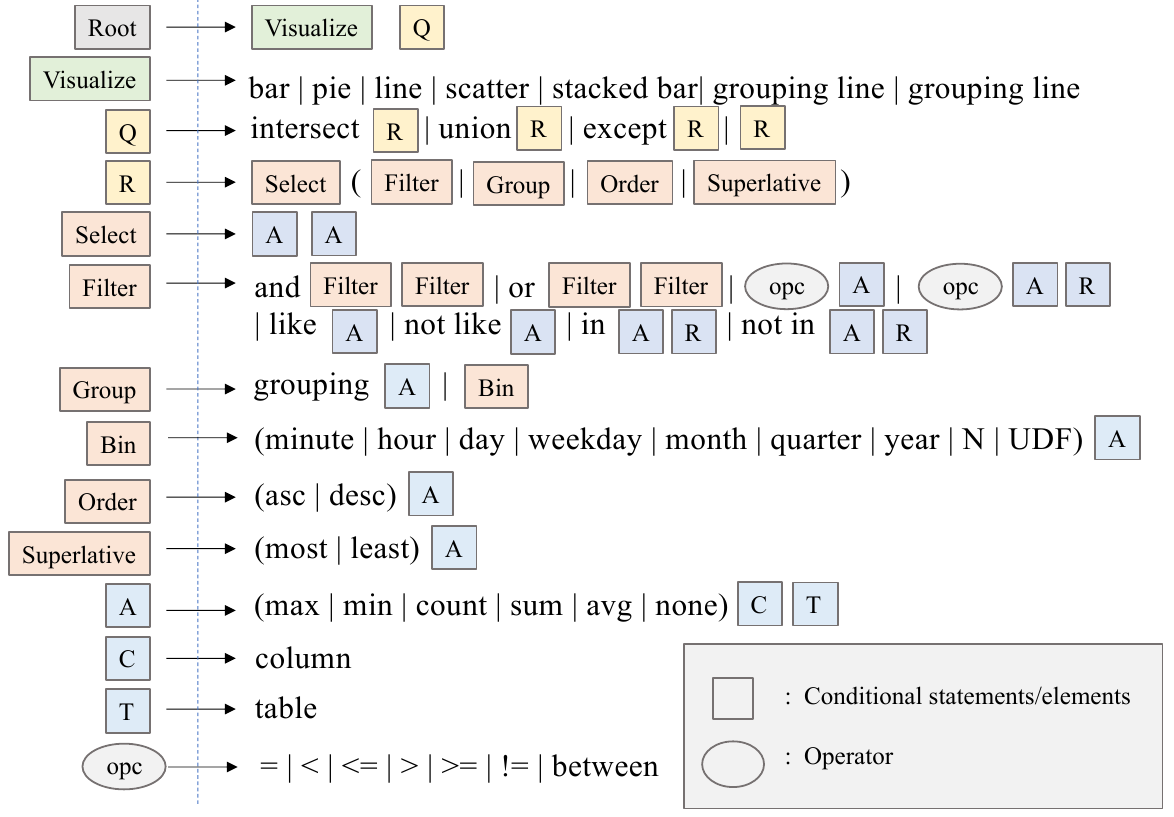}
    \caption{Grammar used to decode SQL-form response.}
    \label{fig:ast}
    \vspace{-15pt}
\end{figure}

\subsubsection{SQL-form Decoder} An important kind of query from the users is data-related, and the essential way of answering them is through an SQL query.  
Due to the strict grammar that the SQL-form response has, which differs from the textual response, we design an SQL grammar-aware decoder here. 
Inspired by the common practice in a paralleled task named \emph{text-to-SQL} \cite{iacob2020neural}, we also employ an intermedia SemQL grammar \cite{guo2019towards} to represent each SQL query. 
The grammar of SemQL is represented in Figure~\ref{fig:ast}, where each row in the figure corresponds to an \textit{action}, and the process of generating SemQL (which could be represented by an Abstract Structure Tree (AST) $\tilde{r}$) is achieved by selecting a series of actions. 

In particular, each step in the action selection process can be regarded as a classification problem, where the candidates and the final choice are determinied by the inputs and outputs of the previous step, represented mathematically as follows. 
\vspace{-5pt}
\begin{equation}
    p(\tilde{r}|q,D,S) = \prod_{i=1}^{T}p(a_i|D,S,q,a_{<i}),
\label{equ:ast-gen}
\end{equation}
where $a_i$ refers to an action performed at step $i$, $a_{<i}$ includes all the previous actions of step $i$, and $T$ indicates the total number of actions needed to obtain the desired SemQL $\tilde{r}$.
The actions are further divided into two types: \emph{ApplyRule} and \emph{SelectItem}. 
The former generates the \textit{sketch} of the AST, the latter selects column and table items to finalize the sketch, and \textit{finally}, we obtain a complete SemSQL AST. 
The mathematical details of ApplyRule can be denoted as follows:
\vspace{-5pt}
\begin{equation}
\begin{aligned}
    \boldsymbol{h}_i &= \texttt{LSTM}([\boldsymbol{a}_{i-1} \oplus \boldsymbol{e}_{i-1} \oplus \boldsymbol{v}_{i-1}], \boldsymbol{h}_{i-1}),
\end{aligned}
\end{equation}
\vspace{-5pt}
\begin{equation}
\begin{aligned}
    \boldsymbol{v}_i &= \texttt{Softmax}(\boldsymbol{h}^T_{i}\boldsymbol{W}_{h}\boldsymbol{H}^T_{Q})\boldsymbol{H}_Q,
\end{aligned}
\end{equation}
\vspace{-5pt}
\begin{equation}
\begin{aligned}
    \boldsymbol{u}_i &= \texttt{tanh}(\boldsymbol{W}_{u}[\boldsymbol{h}_i \oplus \boldsymbol{v}_{i}] + \boldsymbol{b}_u),
\end{aligned}
\end{equation}
\vspace{-5pt}
\begin{equation}
\begin{aligned}
    p(\tilde{r}_i=a_i | q, S, D, a_{<i}) &= \texttt{Softmax}(\texttt{tanh}(\boldsymbol{W}_{p}\boldsymbol{u}_{i} + \boldsymbol{b}_p)),
\end{aligned}    
\label{equ:decoder-sketch}
\end{equation}
where $\oplus$ denotes the concatenation operation. 
$\boldsymbol{h}_i$ is the LSTM hidden state of the current step $i$, $\boldsymbol{a}_{i-1} $ is the previous action embedding,  $\boldsymbol{e}_{i-1}$ is the previous action type (i.e., ApplyRule or SelectItem) embedding and $\boldsymbol{v}_{i-1}$ is the previous contextual representation of LSTM.
$\boldsymbol{W}_h \in \mathbb{R}^{d_{m} \times d_{m}}$, $\boldsymbol{W}_u \in \mathbb{R}^{d_{m} \times 2d_{m}}$ and $\boldsymbol{W}_p \in \mathbb{R}^{n_a \times d_{m}}$ ($n_{a}$ is the number of actions associated with the specified grammar) are the trainable weights, $\boldsymbol{b}_u \in \mathbb{R}^{d_{m}}$ and $\boldsymbol{b}_p \in \mathbb{R}^{n_a}$ are trainable biases, and the initial state $\boldsymbol{h}_0$ is calculated by an average-pooling operation on the output $\boldsymbol{H}_Q$ given by the encoder (in Section~\ref{subsec:dataAwareDialogueSessionEncoder}).

Given the sketch, what we need to do next is to select column items. We first calculate an \emph{NLQ-aware representation} for column $c_k$, and then a \emph{pointer network} is used to obtain the selection probability.
The mathematical representation of this progress is shown as follows:
\vspace{-5pt}
\begin{equation}
    \beta_{k,j} = \frac{\boldsymbol{s}^T_{c_k}  \boldsymbol{q}_j }{\left \| \boldsymbol{s}_{c_k} \right \| \left \| \boldsymbol{q}_j \right \|  },
\end{equation}
\begin{equation}
    \tilde{\boldsymbol{s}}_{c_k} = \boldsymbol{s}_{c_k} +  {\textstyle \sum_{j=1}^{|\boldsymbol{Q}|}\beta_{k,j}\boldsymbol{h}^Q_j},
\end{equation}
\begin{equation}
\begin{aligned}
    \gamma_{k,i} = (\tilde{\boldsymbol{s}}_{c_k})^T\boldsymbol{W}_{c}\boldsymbol{u}_{i},
\end{aligned}    
\end{equation}
\vspace{-10pt}
\begin{equation}
\begin{aligned}
    p(\tilde{r}_i=\textup{SelectColumn}(c_k) | D, S, q, a_{<i}) &= \frac{exp(\gamma_{k,i})}{ {\textstyle \sum_{j=1}^{n_c}}exp(\gamma_{j,i}) },
\end{aligned}    
\end{equation}
where $\boldsymbol{s}_{c_k}$ is the embedding for column $c_k$ obtained from Glove embedding \cite{pennington2014glove}, $\boldsymbol{q}_j$ and $\boldsymbol{h}^Q_j$ are the corresponding vectors of the $j$-th question token extracted from $\mathbf{Q}$ and $\boldsymbol{H}_Q$, respectively, and $\boldsymbol{W}_c \in \mathbb{R}^{d_{e} \times d_{m}}$ is the learnable weights. 

\subsubsection{DV-form Decoder}
The ultimate goal of the DV-form decoder aims to show the desired DV chart to the user. 
Rather than directly generating the charts with the neural network, we design the decoder first to synthesize the DV query. The grammar of the DV query is quite similar to the SQL query, and it was proposed and widely used in the text-to-vis task \cite{luo2021synthesizing, song2022rgvisnet, luo2021natural}. 
We use $r_{sql}$ to represent the SQL query predicted in the last turn by the MMCoVisNet model and $r_{dvq}$ to denote the DV query predicted by the MMCoVisNet model. 
If the model synthesizes an accurate DV query $r_{dvq}$, it yields an accurate DV chart, denoted as $c$. 
Specifically, after the DV query is obtained, a pipeline approach shown in Algorithm~\ref{alg:a1} is triggered to generate the final chart. 
A DV query is composed of the \emph{visualization component} and the \emph{data manipulation component}. 
The visualization component of the DV query is generated by a \emph{Transformer-based decoder} in MMCoVisNet, with a shared structure to the textual decoder described in Section~\ref{sec: textual_decoder}. 
Combined with the previous SQL response and the data retrieved by the SQL query, we can synthesize the visualization specifications $c_{vl}$ in Vega-Lite. 

\begin{algorithm}[t!]
\small
  \SetKwInOut{Input}{input}
  \SetKwInOut{Output}{output}
  {\textbf{Input:}} Current query $q$, Dialogue Context $S$, MMCoVisNet Model $M$, SQL Response in the Last Turn $r_{sql}$, Dataset $D$ \\
  {\textbf{Output:}} Visualization Chart $c$ \\
  \Begin{
  Obtain the DV Query from the model output $r_{dvq}$ $\gets$ $M(q, D, S)$; \\
  Execute the SQL query to retrieval data $d$ $\gets$ execute($D$, $r_{sql}$); \\
  Get Vega-Lite specification $c_{vl}$ $\gets$ transform($r_{dvq}$, $d$)
  DV Chart $c$ $\gets$ render($c_{vl}$); \\
  return $c$; \\
  }
\caption{The Pipeline for Obtaining the DV Charts}
\label{alg:a1}
\end{algorithm}

\vspace{-3pt}
\subsection{Model Training}
Since there exists different types of responses in one continuous multi-turn dialogue and a similar natural language question, we make sure that they share one session encoder but have different decoders. To train them simultaneously, different QR pairs appear in the same batch and are divided into mini-batches by the question classifier. The mini-batch of various types is fed into the network, and the decoder parameters are updated sequentially (textual-form decoder, data-form decoder and DV-form decoder). Then, the encoder is updated only once in the final phase. The widely-used cross-entropy loss is used in our experiments. 
For the textual response ($\mathcal{L}_{1}$) and DV-form response ($\mathcal{L}_{3}$), we maximize the probability of each word appearing in the target sequence by the following function:
\vspace{-5pt}
\begin{equation}
\begin{aligned}
    \mathcal{L}_{3}(r, \hat{r}) = \mathcal{L}_{1}(r, \hat{r})=-\frac{1}{N} \sum_{i=1}^{N}\frac{1}{N_i} \sum_{j=1}^{N_i} \log p(r=\hat{r}_{ij}),
\end{aligned}
\label{equ: loss}
\vspace{-5pt}
\end{equation}
where $\hat{r}$ is the predicted response, $r$ is the target response, $N$ is the total number of training samples, and $N_i$ is the length of the response of the $i$-{th} sample. $p(r=\hat{r}_{ij})$ represents the probability of the predicted $j$-{th} token in the $i$-{th} sample being equal to $\hat{r}_{ij}$. 

For the SQL-form response, unlike above mentioned textual and DV-form responses, it works in a two-step generation style. And thus, we train the model by maximizing the log-likelihood of the ground truth action sequences, defined as
\vspace{-5pt}
\begin{equation}
\begin{aligned}
  \mathcal{L}_{2}(r, \hat{r}) &= \max \sum_{(D, S, q, r) \in \mathcal{T}} \left [ \sum_{a_i \in \textup{ApplyRule}} \log p(\tilde{r}_i = a_i | D, S, q, a_{<i}) \right.\\
   & \left. + \sum_{a_i \in \textup{SelectColumn}} \log p(\tilde{r}_i = a_i | D, S, q, a_{<i}) \right ].
\end{aligned}
\label{equ:loss}
\vspace{-5pt}
\end{equation}

\begin{table*}[th!]
\centering
\caption{Performance Comparison.} 
\scalebox{1.0}{
    \begin{tabular}{l|ccc|cc|cccc}
    \toprule
    & \multicolumn{3}{c|}{\textbf{Textual Response}} & \multicolumn{2}{c|}{\textbf{SQL-form Response}} & \multicolumn{4}{c}{\textbf{DV-form Response}} \\ 
    \midrule
    \textbf{Method} & BLEU & ROUGH & METEOR & Sketch Acc. & SQL Acc. & Vis. Acc. & Axis Acc. & Data Acc. & DV Acc.  \\
    \midrule
    Seq2Seq & 0.7437 & 0.8499 & 0.7954 & - & 14.92\% & 100.00\% & 31.47\% & 33.59\% & 18.97\%  \\
    Transformer & 0.7436 & 0.8451 & 0.7953 & - & 46.72\% & 100.00\% & 67.30\% & 60.49\% & 55.80\%  \\
    HRED & 0.7200 & 0.7732 & 0.7206 & -  & 18.72\% & 99.67\% & 39.73\% & 39.73\% & 25.11\% \\
    \textbf{MMCoVisNet} & 0.7381  & 0.8556 & 0.7958 & 87.11\% & 62.37\% & 99.89\% & 73.55\% & 67.75\% & 67.97\% \\
     \bottomrule
  \end{tabular}
  \label{tb:accuracy}
}
\vspace{-3pt}
\end{table*}

\section{Experimental Setup}
\label{sec:setup}

This section aims to provide a detailed performance evaluation of the proposed model. 
We first describe the experimental setup, the evaluation measurements, and the baselines. Then, we compare the performance of our proposed model with these strong baselines. 

\subsection{Models}

Several popular dialogue models and methods were implemented in our experiment to compare the performance. 

\begin{itemize}
    \item {Seq2Seq}: Seq2Seq \cite{bahdanau2015neural} is a fundamental model that treats the dialogue task as a translation that maps from the queries to the responses. 
    \item {Transformer}: Transformer \cite{vaswani2017attention} has shown a dominant performance in many NLP tasks including machine translation \cite{currey2019incorporating}, dialogue system \cite{zhao2020multiple} and speech recognition \cite{zeyer2019comparison}. Like Seq2Seq, we also modify it as a baseline method in our experiment.  
    \item {HRED}: HRED \cite{serban2016building} is a prevalent and representative model in the field of multi-turn dialog systems. 
    \item {MMCoVisNet}: This is our proposed method, which employs a data-aware session encoder as well as advanced adaptive decoders to generate responses in various forms.
\end{itemize}

To guarantee fairness and reproducibility, all the above-mentioned methods were trained on the same training set and evaluated on the same testing set. 
We try our best to optimize the performance of these models to achieve their best performance. 

\subsection{Evaluation Metrics}

Since our task includes three kinds of outputs, i.e., the textual form response, the SQL-form data response and the DV chart response, we employ various metrics to compare the performance. For the textual response, we employ the classical \textit{BLEU} \cite{papineni2002bleu}, \textit{ROUGH} \cite{lin2004rouge} and \textit{METEOR} \cite{banerjee2005meteor} scores in a textual-form dialogue system as our main metrics.
For SQL-form response, we employ the widely-used \emph{Sketch Accuracy} and \emph{Query Matching Accuracy} \cite{yu2018spider} in the text-to-SQL task as the main indicators.
If a model can generate an accurate SQL query, it will also retrieve the correct data. 
Similarly, we also employ \emph{Vis Accuracy}, \emph{Axis Accuracy}, \emph{Data Accuracy} and \emph{DV Accuracy} used in text-to-vis \cite{luo2021synthesizing} to reflect the performance of the generated DV query. 
For all these metrics, including the ones for evaluating the textual, the SQL-form and the DV-form responses, a larger value of each metric means a better result.
A correctly predicted DV query leads to a correctly generated Vega-Lite configuration and the final chart.

\subsection{Implementation Details}
Our models are trained using the Adam optimizer, with a mini-batch size set to 32 for dialogues and a learning rate set to $1 \times 10^{-4}$. 
The dimensionality of word embedding of Glove is 300, the size of vocabulary used in this experiment is 15,894, and each unseen word is initialized by `<unk>' denoting unknown word. 
For the session encoder, we set the number of layers to 1, the number of heads to 4, and the hidden size to 512. Both textual and DV-form decoders constructed by the Transformer have the same parameters as the encoder. In addition, the dimensionality of action embedding and type embedding in the SQL-form decoder are all set to 128 and the hidden size to 512. 

\subsection{Experimental Results}
\label{sec:exp}

\begin{table*}[th!]
\centering
\caption{Ablation Study Results.} 
\scalebox{1.0}{
     \begin{tabular}{l|ccc|cc|cccc}
    \toprule
    & \multicolumn{3}{c|}{\textbf{Textual Response}} & \multicolumn{2}{c|}{\textbf{SQL-form Response}} & \multicolumn{4}{c}{\textbf{DV-form Response}} \\ 
    \midrule
    \textbf{Method} & BLEU & ROUGH & METEOR & Sketch Acc. & SQL Acc. & Vis. Acc. & Axis Acc. &Data Acc. & DV Acc.  \\
    \midrule
    {MMCoVisNet} & 0.7381  & 0.8556 & 0.7958 & 87.11\% & 62.37\% & 99.89\% & 73.55\% & 67.75\% & 67.97\% \\
    { w/o session} & 0.7480 & 0.8594 & 0.8021 & 58.81\% & 33.33\% & 100.00\% & 43.19\% & 17.97\% & 16.52\% \\
    { w/o dataset} & 0.5106 & 0.8059 & 0.5806 & 89.50\% & 61.45\% & 100.00\% & 74.44\% & 66.29\% & 66.63\% \\
    { w/o both} & 0.5050 &  0.7831 & 0.5607 & 58.44\% & 32.78\% & 100.00\% & 40.40\% & 18.08\% & 17.08\% \\
    { w. basic decoder} & 0.7511 & 0.8593 & 0.8057 & - & 45.67\% & 99.89\% & 66.85\% & 61.94\% & 54.35\% \\
    \bottomrule
  \end{tabular}
  \label{tb:ablation}
}
\end{table*}

\subsubsection{Performance Evaluation.}
\label{sec:sql-result}

We compare the performance in the three forms of the response generated in Table~\ref{tb:accuracy}. 
Since all the baseline methods directly generate the SQL query rather than first predicting a sketch, only our proposed MMCoVisNet contains the value in sketch accuracy. 
From the result, we report the following observations. 

Since the data-related queries are critical and significantly affect the accuracy of the following generated DV charts, we first evaluate the performance of SQL-from response generation. The vanilla end-to-end model, Seq2Seq, performs poorly and is not quite competitive as a baseline, due to two factors: (i) its capability is limited in capturing the semantics contained in the dialogue session and the dataset; (ii) its ability is also limited in generating the SQL queries, since SQL enjoys a very strict and complicated grammar. 
Other advanced models, such as Transformer and HRED, could improve the vanilla Seq2Seq model from these two aspects. 
For example, the HRED model employs a hierarchical decoder to generate responses for different inputs, leading to a 32.37\% relative improvement in the DV accuracy. 

Our designed {MMCoVisNet} could make full use of all the information (i.e., dialogue history, query and dataset) out of all possible models, and at the same time, employ adaptive decoders to generate accurate responses. 
Thus, it remarkably beats all the baseline methods, including advanced models such as HRED and Transformer, thus proving the efficacy and validating the necessity of designing an advanced encoder and decoder in the CoVis scenario. 

For the final DV-form responses, the basic end-to-end approach, Seq2Seq, performs poorly and is not quite competitive as a baseline. Other models (e.g., Transformer and HRED) could beat the vanilla Seq2Seq model. 
Even though our proposed MMCoVisNet model is slightly weaker in the metric \emph{Vis. Accuracy} ($\approx$ 0.1\%), it achieves the best performance over all other metrics. 
Most importantly, it largely surpasses all other baselines in the final \emph{DV Accuracy} (e.g., a 21.8\% relative improvement over the Transformer). 
We also notice that the accuracy of the DV-form response is much higher than the SQL-form query. 
The reason is also apparent. The proposed Dial-NVBench dataset contains more SQL-related queries than the DV-related queries, and a large portion of these SQL-related queries are complicated queries (e.g., nested operations), resulting to an overall 62.37\% SQL accuracy by MMCoVisNet. Among all these SQL queries, only parts of them are being derived into the final DV and this part is relatively easier, leading to a higher SQL query accuracy compared with the \emph{overall} SQL query accuracy and thus a higher DV accuracy (i.e., 67.97\%).

Finally, we also analyze the quality of the textual form responses generated by various methods. 
Since this dataset is mainly for DV generation, the textual responses generation part in this dataset mainly focuses on talking about the dataset and not very diverse. 
And thus, all these models could achieve relative good results. For some metric like BLEU, the difference between these method is slight (e.g., $<$ 0.76\% between Seq2Seq, Transformer and MMCoVisNet).
From the result, we could see that our model could still generate comparable results even if baselines such as HRED are dedicated to and specially designed for textual dialogue generation. The main reason for this is that our model can understand the dialogue context, which leads to accurate textual responses.

\subsubsection{Ablation Studies}
\label{sec:Quantitative Comparison}

This section shows the ablation studies to evaluate the effectiveness and contribution of each designed component in MMCoVisNet. 
Firstly, we examine the performance of the MMCoVisNet model with all its proposed components.
Then, we investigate the performance of the MMCoVisNet model by removing or replacing its components. 
We first check each part in the encoder. To evaluate the effectiveness of the dialogue session component, we removed it and named this baseline \textbf{w/o session}. Similarly, to analyze the effectiveness of the dataset, we removed it and called it the \textbf{w/o dataset}. We also remove both the dialogue session and the dataset components, and we named it \textbf{w/o both}.
For the decoder part, we replace our designed adaptive decoder with a vanilla LSTM-based one (\textbf{w. basic decoder}). The results are shown in Table~\ref{tb:ablation}. 

We take the DV accuracy as the primary indicator, and the other metrics reflect similar conclusions. 
First, integrating the session information leads to {51.45\%} absolute performance improvement ({67.97\% v.s 16.52\%}). 
This ablation study set and its substantial improvement demonstrate the necessity of incorporating dialogue session information in this CoVis task. Other components, like involving dataset information, show similar observations. 
Then, for the decoder part, using a basic LSTM-based decoder directly generated the SQL query rather than a sketch first, and it has no value in sketch accuracy. 
Compared with a basic LSTM-based model, our designed adaptive decoder shows about {16.37\%} relative improvement, proving the necessity of using advanced NN-based models in generating diverse responses.

\begin{figure}[th!]
\vspace{-5pt}
  \centering
    \subfigure[{Textual-form Perf.}]{
   \centering
     \includegraphics[width=0.145\textwidth]{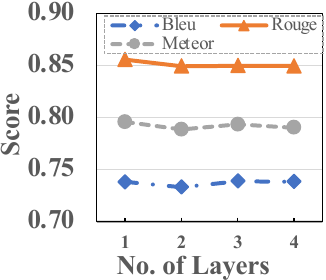}
     \label{fig:proto-num-val}
  }
   \subfigure[{SQL-form Perf.}]{
   \centering
     \includegraphics[width=0.145\textwidth]{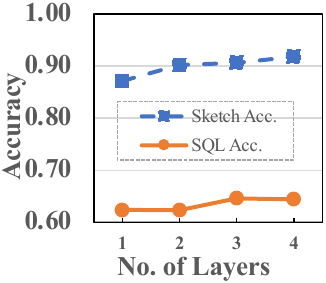}
     \label{fig:gnn-layer-val}
   }
  \subfigure[{DV-form Perf.}]{
   \centering
     \includegraphics[width=0.145\textwidth]{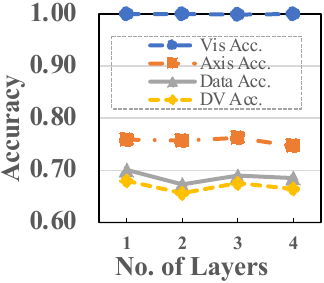}
     \label{fig:dv-layer}
  }
  \vspace{-5pt}
  \caption{Vary No. of Layers in Encoder}
\label{fig:parameters-layer}
\vspace{-10pt}
\end{figure}
\begin{figure}[ht!]
\vspace{-10pt}
  \centering
    \subfigure[{Textual-form Perf.}]{
   \centering
     \includegraphics[width=0.15\textwidth]{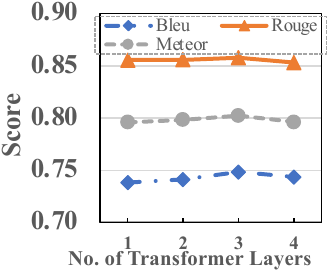}
     \label{fig:proto-num-val}
  }
   \subfigure[{SQL-form Perf.}]{
   \centering
     \includegraphics[width=0.145\textwidth]{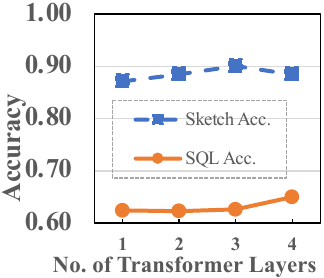}
     \label{fig:gnn-layer-val}
   }
  \subfigure[{DV-form Perf.}]{
   \centering
     \includegraphics[width=0.145\textwidth]{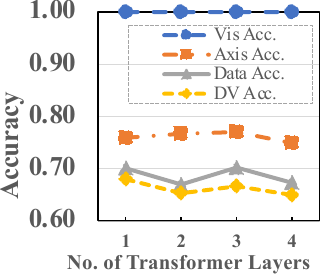}
     \label{fig:tran-num-val}
  }
  \caption{Vary No. of Transformer Layers}
    \label{fig:parameters}
    \vspace{-10pt}
\end{figure}
\begin{figure}[ht!]
\vspace{-10pt}
  \centering
    \subfigure[{Textual-form Perf.}]{
   \centering
     \includegraphics[width=0.145\textwidth]{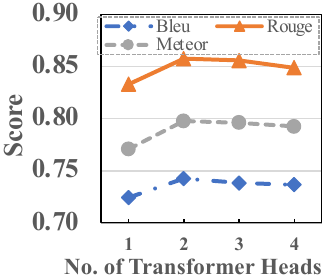}
     \label{fig:proto-num-val}
  }
   \subfigure[{SQL-form Perf.}]{
   \centering
     \includegraphics[width=0.145\textwidth]{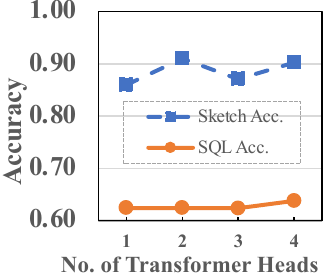}
     \label{fig:gnn-layer-val}
   }
  \subfigure[{DV-form Perf.}]{
   \centering
     \includegraphics[width=0.145\textwidth]{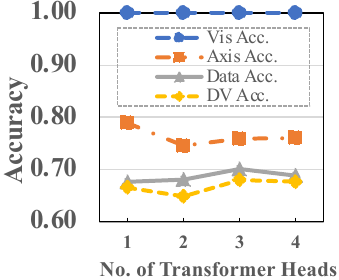}
     \label{fig:tran-num-val}
  }
  \caption{Vary No. of Transformer Heads}
\label{fig:parameters-head}
\vspace{-10pt}
\end{figure}

\begin{table*}[th!]
\includegraphics[width=1.0\textwidth]{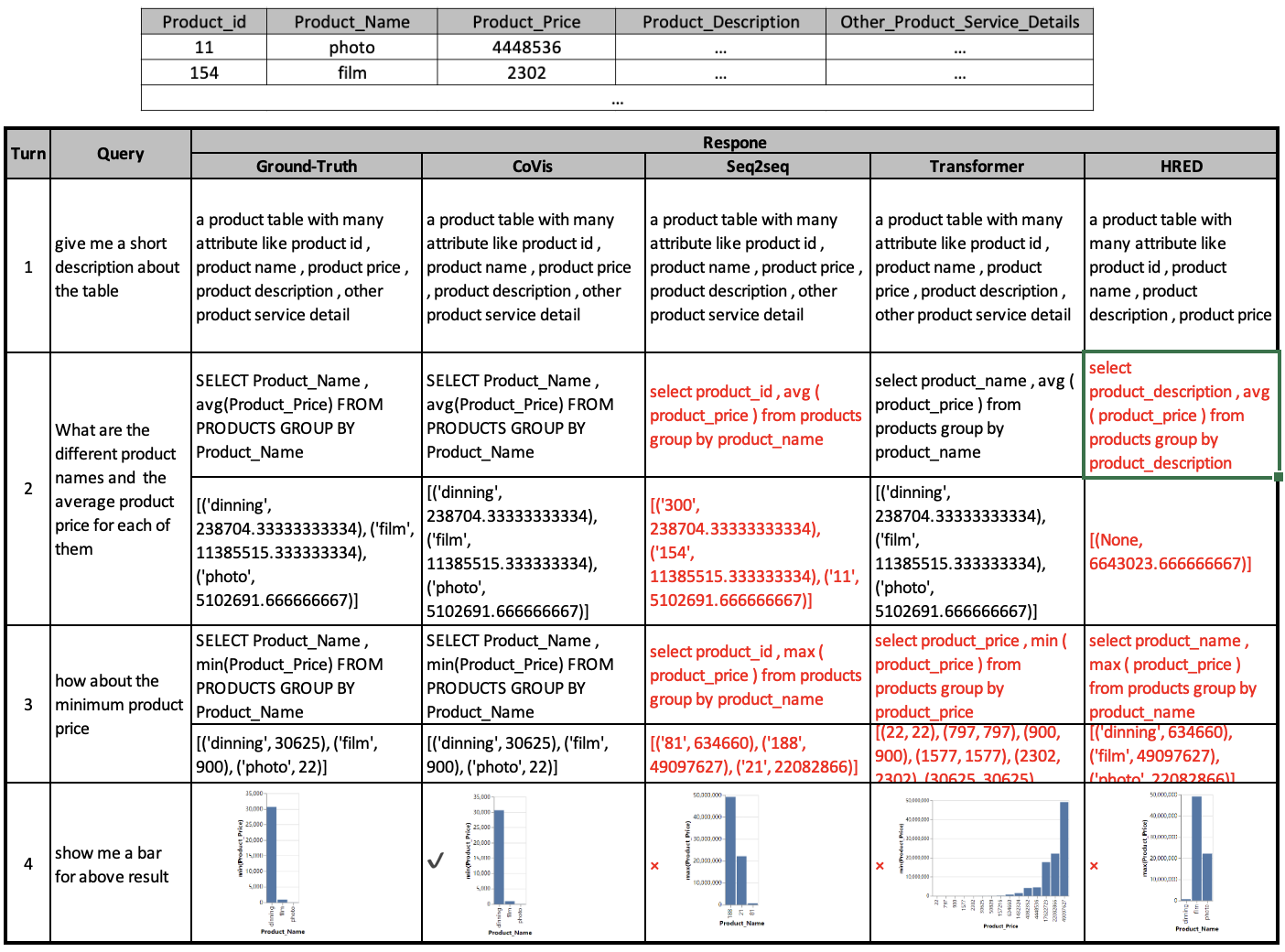}
\caption{Examples of Queries \& Responses Returned by Various CoVis Methods (errors are marked with red color and $\times$).}
\label{tab:case}
\vspace{-10pt}
\end{table*}

\subsection{Hyper-parameter Study}
\label{sec:hyper study}
This set of experiments analyze the influence of parameter variations on the performance of our proposed MMCoVisNet model. 
We first identify three main parameters that affect the performance of the model, namely the number of layers in the encoder (Figure~\ref{fig:parameters-layer}), the number of heads (Figure~\ref{fig:parameters}), and the number of layers in the Transformer (Figure~\ref{fig:parameters-head}). We change the value of one parameter and fix the remaining ones each time. 

We utilize the DV accuracy in Figure~\ref{fig:dv-layer} as the main indicator in the following analysis. 
The accuracies achieve the best when the number of layers in the encoder is set to 3, since they decrease when the number of layers overpasses or underpasses this optimal value. 
In particular, we observe that increasing this value does not always result in a better performance. 
This phenomenon is also observed in previous studies like \cite{li2019deepgcns}, since too many layers usually result in the smoothing problem.
For other parameters, i.e., the number of Transformer layers (Figure~\ref{fig:parameters}) and the number of Transformer heads (Figure~\ref{fig:parameters-head}), we also observe similar performance trends. 

\subsection{Case Study}
\label{sec:case study}

We also list a real case shown in Table~\ref{tab:case} to give a vivid illustration, which includes the original datasets to be analyzed, the queries and the predicted responses (text, SQL query \& data, and DV charts) by various methods. 
In the example, the user directly queries the details about a ``Product'' dataset by asking general question like \emph{``give me a short description about the table''}, data-related question like \emph{``what are the different product names and the average product price for each of them''}, and DV-related question like \emph{``show me a bar for the above result''}. From this example, we could see that MMCoVisNet could accurately understand the meaning of all the queries and generate correct answers for text, data and DV. Other baselines, including the advanced Transformer and HRED, fail in answering most of the queries and also obtained incorrect DV charts finally. 
It is noted that MMCoVisNet also makes full use of the meaning in the whole dialogue session rather than a solely current query, which is very important for understanding queries like \emph{``how about the minimum product price''} (Turn 3) that refers to the previous context (Turn 2).

\section{Related Work}
\label{sec:related}
This work is related to the research areas of data visualization, natural languages to SQL, and dialogue systems. 
This section briefly surveys the most related work from these three fields. 

\subsection{Data Visualization} 
Recent years have witnessed an enormous rise in DV in the data mining \cite{song2022rgvisnet,qian2021learning,ho2002visualization,fayyad2002information} and the database communities \cite{tang2022sevi,hanrahan2006vizql,luo2021synthesizing,vartak2017towards,luo2018deepeye2}. 
There are many intelligent tasks proposed to lower the barriers to use DV. We want to cover two typical ones: text-to-vis and DV recommendation. 

Text-to-vis focuses on automatically translating a natural language question (NLQ) into its corresponding DV. It also enables non-experts to manipulate the DV system easily.
The most popular method right now is to treat it as a machine translation problem and then utilize deep neural networks to map from the input (i.e., NLQ) to the output (i.e., DV). For example, Cui \textit{et al.} introduces text-to-viz and utilizes rule-based methods to translate text statements to infographics \cite{cui2019text}. 
Draco-Learn \cite{moritz2018formalizing} views the visualization design as a selection process from a set of constraints and then use experimental data to turn the weights for these soft constraints.
Data2Vis \cite{dibia2019data2vis} also attempts to convert the data series to visualization specifications in some declarative languages in a machine translation style.
NL4DV \cite{narechania2020nl4dv} is a Python toolkit that offers a number of high-level operations to assist users in building DV systems with NL-interfaces (NLIs).
To alleviate the problem of data scarcity in this field, Luo \textit{et al.} additionally describes an approach to synthesize the NLQ-DV dataset called NVBench based on the popular NL2SQL benchmark Spider \cite{yu2018spider}. 
A Seq2Seq model is further trained on this dataset to validate the feasibility of DV query generation from NLQs on this benchmark \cite{luo2021synthesizing}. RGVisNet \cite{song2022rgvisnet} is another representative study in KDD'22 that employs a DNN-based approach for converting NLQs into DVs. However, it beats other methods by combining the retrieval-based method with the generative-based one. 
In contrast, automatic DV recommendation directly outputs top-$N$ potential DVs given a dataset without any NL queries from the users. 
To name a few, 
DataEye \cite{luo2018deepeye} tackles the DV recommendation problem in a three-step style, namely visualization recognition, ranking and selection. 
\cite{qian2021learning} in KDD'21 employs an end-to-end learning-based method to construct DVs, giving a massive dataset. 

\smallskip
Different from these studies, we propose a novel DV-related task named CoVis (i.e., Conversational text-to-Vis), which focuses on \emph{interactive exchange} between the user and the DV system. With the benchmark and all the baseline methods proposed in this paper, CoVis would become a prevalent DV-related task and inspire more studies in \textit{NLP for Data Mining} direction. 

\subsection{Natural Language to SQL}
Natural language to SQL (NL2SQL or text-to-SQL) is another closely related area that motivates text-to-vis development. 
Its ultimate goal is to enable databases with NL-based interfaces (NLIs). 
Past literature in this field first handles the \emph{single-turn} text-to-SQL translation with representative studies such as Seq2SQL \cite{data-wikisql}, SQLNet \cite{xu2017sqlnet}, TypeSQL \cite{yu2018typesql}, Syntax SQL \cite{sun2018semantic}, IRNet \cite{guo2019towards}. \cite{affolter2019comparative} and \cite{iacob2020neural} are two representative surveys about these studies. Moreover, numerous open-source text-to-SQL benchmarks have been released to promote research in this field, such as Spider \cite{yu2018spider} and WikiSQL \cite{data-wikisql}. 
Similarly, researchers also extend the vanilla text-to-SQL task to a multi-turn scenario with a public dataset such as SparC \cite{yu2019sparc} and CoSQL \cite{yu2019cosql}. The multi-turn interaction between the user and the NLI for DB systems would greatly improve the accuracy and further reduce the translation errors when converting NLQs to SQL queries, because it provides more chances to correct the errors in a sequence of conversations.

\smallskip
The success of the dialogue system and CoSQL also validates the necessity of \emph{multi-turn exchanges} between machines and users in database scenarios. Our work is the first to study multi-turn exchanges between machines and users in the DV area. 

\subsection{Dialogue System}
Tremendous efforts have been made by the community in designing various dialogue systems. 
The existing dialogue systems could be roughly divided into \emph{open-domain} and \emph{task-oriented ones}. 
The former focuses on providing chitchat or general responses to the users with typical systems like Microsoft's XiaoIce \cite{zhou2020design}, Alibaba's AliMe \cite{li2017alime}, Google's Meena \cite{meena,adiwardana2020towards}, and Meta's BlenderBot \cite{blender}. 
Regarding implementation, retrieval-based and generation-based approaches were proposed for this task. 
The retrieval-based approach usually \emph{chooses} the most appropriate instances from a library as the responses, whereas the generation-based method employs learning-based models to \emph{generate} responses \cite{jokinen2009spoken}. 
Recently, studies on combining both approaches were conducted to enjoy the merits of both methods \cite{song2018ensemble,yang2019hybrid}. 
In contrast, the objective of task-oriented dialogue systems is typically to assist users to complete specific tasks in specific domains \cite{chen2018dialogue,zhang2020recent}.
Typical approaches to the task-oriented dialogue systems includes hybrid solutions \cite{balaraman2021recent} and the end-to-end ones \cite{li2017end,wen2017network,yang2021ubar}.
In addition to textual dialogue systems, multi-modal dialogue systems are another direction that shows excellent usefulness in scenarios such as product recommendation in e-commerce \cite{saha2018towards,liao2018knowledge,he2020multimodal,nie2019multimodal}. However, these multi-modal dialogue systems (e.g., MAGIC \cite{nie2019multimodal}) usually treat the product recommendation task as an image selection one from a candidate pool. 
Our task requires the model to \emph{generate the responses from scratch} without any image candidate pool, which is much harder than selecting appropriate images from a large set of candidates. Thus, these exsiting multi-modal dialogue systems cannot perform well on this new proposed CoVis task.

\smallskip
Our proposed CoVis task marries the popular DV task with dialogue system, and the proposed MMCoVisNet model enlarges the family of multi-modal dialogue systems and would greatly alleviate the barriers to using DV systems for beginners and non-technical background users.  

\section{Conclusion}
\label{sec:con}
We introduce a new task named CoVis, aiming to generate data visualizations from conversations, and it could be used to construct interactive DV dialogue systems. 
To promote this field, we first construct a benchmark dataset, which includes a sequence of dialogues with diverse queries and responses. 
The ultimate goal of each dialogue session is to create a suitable DV. However, this process contains various dialogue queries, like querying the information about the dataset, querying parts of the data, and visualizing the data. Then, we propose a multi-modal neural network named MMCoVisNet to answer these DV-related queries.
Experimental results have validated that MMCoVisNet achieves superior performance over other existing baselines. 

This study is a good start in this new CoVis task. 
After we validate the feasibility of this task, there are many potential studies following this line of research. 
For example, a potential direction would be designing advanced neural structures like graph neural networks for modeling the connections in the dialogue context. 

\bibliographystyle{ACM-Reference-Format}
\bibliography{ref}

\end{document}